# Learning Module Networks

**Eran Segal**

Computer Science Dept.
Stanford University
Stanford, CA 94305, USA
eran@cs.stanford.edu

**Dana Pe'er**

Computer Science & Eng.
Hebrew University
Jerusalem, 91904, Israel
danab@cs.huji.ac.il

**Aviv Regev**

Bauer Ctr Genomic Research
Harvard University
Cambridge, MA 02138, USA
ARegev@CGR.harvard.edu

**Daphne Koller**

Computer Science Dept.
Stanford University
Stanford, CA 94305, USA
koller@cs.stanford.edu

**Nir Friedman**

Computer Science & Eng.
Hebrew University
Jerusalem, 91904, Israel
nir@cs.huji.ac.il

## Abstract

Methods for learning Bayesian networks can discover dependency structure between observed variables. Although these methods are useful in many applications, they run into computational and statistical problems in domains that involve a large number of variables. In this paper, we consider a solution that is applicable when many variables have similar behavior. We introduce a new class of models, *module networks*, that explicitly partition the variables into modules that share the same parents in the network and the same conditional probability distribution. We define the semantics of module networks, and describe an algorithm that learns the modules' composition and their dependency structure from data. Evaluation on real data in the domains of gene expression and the stock market shows that module networks generalize better than Bayesian networks, and that the learned module network structure reveals regularities that are obscured in learned Bayesian networks.

## 1 Introduction

Over the last decade, there has been much research on the problem of learning Bayesian networks from data [13], and successfully applying it both to density estimation, and to discovering dependency structures among variables. Many real-world domains, however, are very complex, involving thousands of relevant variables. Examples include modeling the dependencies among expression levels ($\approx$ activity) of all the genes in a cell [10, 17] or among changes in stock prices. Unfortunately, in complex domains, the amount of data is rarely enough to robustly learn a model of the underlying distribution. In the gene expression domain, a typical data set includes thousands of variables, but at most a few hundred instances. In such situations, statistical noise is likely to lead to spurious dependencies, resulting in models that significantly overfit the data.

In this paper, we propose an approach to address this issue. We start by observing that, in many large domains, the variables can be partitioned into sets so that, to a first approximation, the variables within each set have a similar set of dependencies and therefore exhibit a similar behavior. For example, many genes in a cell are organized into

*modules*, in which sets of genes required for the same biological function or response are co-regulated by the same inputs in order to coordinate their joint activity. As another example, when reasoning about thousands of NASDAQ stocks, entire sectors of stocks often respond together to sector-influencing factors (e.g., oil stocks tend to respond similarly to a war in Iraq).

We define a new representation called a *module network*, which explicitly partitions the variables into *modules*. Each module represents a set of variables that have the same statistical behavior, i.e., they share the same set of parents and local probabilistic model. By enforcing this constraint on the learned network, we significantly reduce the complexity of our model space as well as the number of parameters. These reductions lead to to more robust estimation and better generalization on unseen data.

A module network can be viewed simply as a Bayesian network in which variables in the same module share parents and parameters. Indeed, probabilistic models with shared parameters are common in a variety of applications, and are also used in other general representation languages, such as *dynamic Bayesian networks* [6], *object-oriented Bayesian Networks* [15], and *probabilistic relational models* [16, 8]. (See Section 7 for further discussion of the relationship between module networks and these formalisms.) In most cases, the shared structure is imposed by the designer of the model, using prior knowledge about the domain. A key contribution of this paper is the design of a learning algorithm that directly searches for and finds sets of variables with similar behavior, which are then defined to be a module. Noise in the data makes it extremely unlikely that such a modular structure would arise naturally from a Bayesian network learning algorithm, even if it exists in the domain. Moreover, by making the modular structure explicit, the module network representation provides insight about the domain that are often be obscured by the intricate details of a large Bayesian network structure.

We describe the basic semantics of the module network framework, present a Bayesian scoring function for module networks, and provide an algorithm that learns both the assignment of variables to modules and the probabilis-tic model for each module. We evaluate the performance of our algorithm on two real datasets, in the domains of gene expression and the stock market. Our results show that our learned module network generalizes to unseen test data much better than a Bayesian network. They also illustrate the ability of the learned module network to reveal high-level structure that provides important insights.

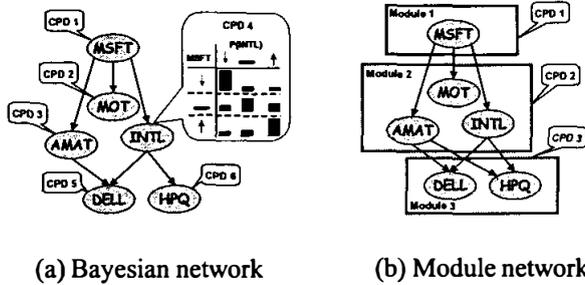

(a) Bayesian network        (b) Module network

Figure 1: (a) A simple Bayesian network over stock price variables; the stock price of Intel (*INTL*) is annotated with a visualization of its CPD, described as a different multinomial distribution for each value of its influencing stock price Microsoft (*MSFT*). (b) A simple module network; the boxes illustrate modules, where stock price variables share CPDs and parameters.

## 2 The Module Network Framework

We start with an example that introduces the main idea of module networks and then provide a formal definition. For concreteness, consider a simple toy example of modeling changes in stock prices. The Bayesian network of Figure 1(a) describes dependencies between different stocks. In this network each random variable corresponds to the change in price of a single stock. For simplicity these random variables take one of three values: 'down', 'same' or 'up', denoting the change during a particular trading day. In our example, the stock price of Intel (*INTL*) depends on that of Microsoft (*MSFT*). The CPD shown in the figure indicates that the behavior of Intel's stock is similar to that of Microsoft. That is, if Microsoft's stock goes up, there is a high probability that Intel's stock will also go up and vice versa. Similarly, the Bayesian network specifies a CPD for each stock price as a stochastic function of its parents. Thus, in our example, the network specifies a separate behavior for each stock.

The stock domain, however, has higher order structural features that are not explicitly modeled by the Bayesian network. For instance, we can see that the stock price of Microsoft (*MSFT*) influences the stock price of all of the major chip makers — Intel (*INTL*), Applied Materials (*AMAT*), and Motorola (*MOT*). In turn, the stock price of computer makers Dell (*DELL*) and Hewlett Packard (*HPQ*), are influenced by the stock prices of their chip suppliers — Intel and Applied Materials. To a first approximation, we can say that the stock price of all chip making companies depends on that of Microsoft and in much the same way. Similarly, the stock price of computer makers that buy their chips from Intel and Applied Materials depends on these chip makers' stock and in much the same way.

To model this type of situation, we might divide stock price variables into groups, which we call *modules*, and require that variables in the same module have the same probabilistic model; that is, all variables in the module have the same set of parents and the same CPD. Our example contains three modules: one containing only Microsoft, a second containing chip makers Intel, Applied Materials, and Motorola, and a third containing computer makers Dell and HP (see Figure 1(b)). In this model, we need only specify three CPDs, one for each module, since all the variables in each module share the same CPD. By comparison, six different CPDs are required for a Bayesian network representation. This notion of a module is the key idea underlying the module network formalism.

We now provide a formal definition a module network. Throughout this paper, we assume that we are given a domain of random variables $\mathcal{X} = \{X_1, \ldots, X_n\}$. We use $Val(X_i)$ to denote the domain of values of the variable $X_i$.

As described above, a module represents a set of variables that share the same set of parents and the same CPD. As a notation, we represent each module by a *formal variable* that we use as a placeholder for the variables in the module. A *module set* $\mathcal{C}$ is a set of such formal variables $\mathbf{M}_1, \ldots, \mathbf{M}_K$. As all the variables in a module share the same CPD, they must have the same domain of values. We represent by $Val(()\mathbf{M}_j)$ the set of possible values of the formal variable of the $j$'th module.

A module network relative to $\mathcal{C}$ consists of two components. The first defines a template probabilistic model for each module in $\mathcal{C}$; all of the variables assigned to the module will share this probabilistic model.

**Definition 2.1:** A *module network template* $\mathcal{T} = (\mathcal{S}, \theta)$ for $\mathcal{C}$ defines, for each module $\mathbf{M}_j \in \mathcal{C}$:

- a set of parents $\mathbf{Pa}_{\mathbf{M}_j} \subset \mathcal{X}$;
- a *conditional probability template (CPT)* $P(\mathbf{M}_j \mid \mathbf{Pa}_{\mathbf{M}_j})$ which specifies a distribution over $Val(\mathbf{M}_j)$ for each assignment in $Val(\mathbf{Pa}_{\mathbf{M}_j})$.

We use $\mathcal{S}$ to denote the dependency structure encoded by $\{\mathbf{Pa}_{\mathbf{M}_j} : \mathbf{M}_j \in \mathcal{C}\}$ and $\theta$ to denote the parameters required for the CPTs $\{P(\mathbf{M}_j \mid \mathbf{Pa}_{\mathbf{M}_j}) : \mathbf{M}_j \in \mathcal{C}\}$. ∎

In our example, we have three modules $M_1$, $M_2$, and $M_3$, with $\mathbf{Pa}_{\mathbf{M}_1} = \emptyset$, $\mathbf{Pa}_{\mathbf{M}_2} = \{MSFT\}$, and $\mathbf{Pa}_{\mathbf{M}_3} = \{AMAT, INTL\}$.

The second component is a module assignment function, that assigns each variable $X_i \in \mathcal{X}$ to one of the $K$ modules, $\mathbf{M}_1, \ldots, \mathbf{M}_K$. Clearly, we can only assign a variable to a module that has the same domain.

**Definition 2.2:** A *module assignment function* for $\mathcal{C}$ is a function $\mathcal{A} : \mathcal{X} \to \{1, \ldots, K\}$ such that $\mathcal{A}(X_i) = j$ only if $Val(X_i) = Val(\mathbf{M}_j)$. ∎

In our example, we have that $\mathcal{A}(MSFT) = 1$, $\mathcal{A}(MOT) = 2$, $\mathcal{A}(INTL) = 2$, and so on.

A module network is defined by both the module network template and the assignment function.

**Definition 2.3:** Let $\mathcal{M}$ be a triple $(\mathcal{C}, \mathcal{T}, \mathcal{A})$, where $\mathcal{C}$ is a module set, $\mathcal{T}$ is a module network template for $\mathcal{C}$, and $\mathcal{A}$ is a module assignment function for $\mathcal{C}$. $\mathcal{M}$ defines a directed *module graph* $\mathcal{G}_{\mathcal{M}}$ as follows:

- the nodes in $\mathcal{G}_{\mathcal{M}}$ correspond to the modules in $\mathcal{C}$;
- $\mathcal{G}_{\mathcal{M}}$ contains an edge $\mathbf{M}_j \to \mathbf{M}_k$ if and only if there is a variable $X \in \mathcal{X}$ so that $\mathcal{A}(X) = j$ and $X \in \mathbf{Pa}_{\mathbf{M}_k}$.

We say that $\mathcal{M}$ is a a *module network* if the module graph $\mathcal{G}_{\mathcal{M}}$ is acyclic. ∎

For example, for the module network of Figure 1(b), the module graph has the structure $\mathbf{M}_1 \to \mathbf{M}_2 \to \mathbf{M}_3$.

A module network defines a probabilistic model by using the formal random variables $\mathbf{M}_j$ and their associated CPTs as templates that encode the behavior of all of the variables assigned to that module. Specifically, we define the semantics of a module network by "unrolling" a Bayesian network where all of the variables assigned to module $\mathbf{M}_j$ share the parents and conditional probability template assigned to $\mathbf{M}_j$ in $\mathcal{T}$.

**Definition 2.4:** A module network $\mathcal{M} = (\mathcal{C}, \mathcal{T}, \mathcal{A})$ defines a *ground Bayesian network* $\mathcal{B}_{\mathcal{M}}$ over $\mathcal{X}$ as follows: For each variable $X_i \in \mathcal{X}$, where $\mathcal{A}(X_i) = j$, we define the parents of $X_i$ in $\mathcal{B}_{\mathcal{M}}$ to be $\mathbf{Pa}_{\mathbf{M}_j}$, and its conditional probability distribution to be $P(\mathbf{M}_j \mid \mathbf{Pa}_{\mathbf{M}_j})$, as specified in $\mathcal{T}$. The distribution associated with $\mathcal{M}$ is the one represented by the Bayesian network $\mathcal{B}_{\mathcal{M}}$. ∎

Returning to our example, the Bayesian network of Figure 1(a) is the ground Bayesian network of the module network of Figure 1(b).

To show that the semantics for a module network is well-defined, we need to prove that the ground Bayesian network defines a coherent probabilistic model. We need only show the following result:

**Proposition 2.5:** *If $\mathcal{G}_{\mathcal{M}}$ is a directed acyclic graph, then the dependency graph of $\mathcal{B}_{\mathcal{M}}$ is acyclic.*

**Corollary 2.6:** *For any module network $\mathcal{M}$, $\mathcal{B}_{\mathcal{M}}$ defines a coherent probability distribution over $\mathcal{X}$.*

As we can see, a module network provide a succinct representation of the ground Bayesian network. In a realistic version of our stock example, we might have several thousands of stocks. A Bayesian network in this domain needs to represent thousands of CPDs. On the other hand, a module network can represent a good approximation of the domain using a model that uses only few dozen CPDs.

## 3 Bayesian Scoring

We now turn to the task of learning module networks from data. We are given a training set $\mathcal{D} = \{\mathbf{x}[1], \ldots, \mathbf{x}[M]\}$, consisting of $M$ instances drawn independently from an unknown distribution $P(\mathcal{X})$. We assume that the set of modules $\mathcal{C}$ is given, and we wish to estimate this distribution using a module network over $\mathcal{C}$. To provide a complete description of a module network as in Definition 2.3, we need to learn the assignment function $\mathcal{A}$ of nodes to modules, the parent structure $\mathcal{S}$ specified in $\mathcal{T}$, and the parameters $\theta$ for the local probability distributions $P(\mathbf{M}_j \mid \mathbf{Pa}_{\mathbf{M}_j})$. For the remainder of this discussion, we omit references to $\mathcal{C}$, taking it as given.

We take a *score-based approach* to learning module networks. In this section, we define a scoring function that measures how well each candidate model fits the observed data. We adopt the Bayesian philosophy and derive a Bayesian scoring function similar to the Bayesian score for Bayesian networks [5, 14]. In the next section, we consider how to find a high scoring model.

### 3.1 Likelihood Function

We start by examining the *data likelihood* function

$$L(\mathcal{M} : \mathcal{D}) = P(\mathcal{D} \mid \mathcal{M}) = \prod_{m=1}^{M} P(\mathbf{x}[m] \mid \mathcal{T}, \mathcal{A}).$$

This function plays a key role both in the parameter estimation task and in the definition of the structure score.

As the semantics of a module network is defined via the ground Bayesian network, we have that, in the case of complete data, the likelihood decomposes into a product of *local likelihood functions*, one for each variable. In our setting, however, we have the additional property that the variables in a module share the same local probabilistic model. Hence, we can aggregate these local likelihoods, obtaining a decomposition according to modules.

More precisely, let $\mathbf{X}^j = \{X \in \mathcal{X} \mid \mathcal{A}(X) = j\}$, and let $\theta_{\mathbf{M}_j \mid \mathbf{Pa}_{\mathbf{M}_j}}$ be the parameters associated with the CPT $P(\mathbf{M}_j \mid \mathbf{Pa}_{\mathbf{M}_j})$. We can decompose the likelihood function as a product of *module likelihoods*, each of which can be calculated independently and depends only on the values of $\mathbf{X}^j$ and $\mathbf{Pa}_{\mathbf{M}_j}$, and on the parameters $\theta_{\mathbf{M}_j \mid \mathbf{Pa}_{\mathbf{M}_j}}$:

$$
\begin{aligned}
&L(\mathcal{M} : \mathcal{D}) \\
&= \prod_{j=1}^{K} \left[ \prod_{m=1}^{M} \prod_{X_i \in \mathbf{X}^j} P(x_i[m] \mid \mathbf{pa}_{\mathbf{M}_j}[m], \theta_{\mathbf{M}_j \mathbf{Pa}_{\mathbf{M}_j}}) \right] \\
&= \prod_{j=1}^{K} L_j(\mathbf{Pa}_{\mathbf{M}_j}, \mathbf{X}^j, \theta_{\mathbf{M}_j \mid \mathbf{Pa}_{\mathbf{M}_j}} : \mathcal{D}) \qquad (1)
\end{aligned}
$$

If we are learning conditional probability distributions from the exponential family (e.g., discrete distribution,

Gaussian distributions, and many others), then the local likelihood functions can be reformulated in terms of *sufficient statistics* of the data. The sufficient statistics summarize the relevant aspects of the data. Their use here is similar to that in Bayesian networks [13], with one key difference. In a module network, all of the variables in the same module share the same parameters. Thus, we pool all of the data from the variables in $\mathbf{X}^j$, and calculate our statistics based on this pooled data. More precisely, let $S_j(M_j, \mathbf{U})$ be a sufficient statistic function for the CPT $P(M_j \mid \mathbf{U})$. Then the value of the statistic on the data set $\mathcal{D}$ is

$$\hat{S}_j = \sum_{m=1}^{M} \sum_{X_i \in \mathbf{X}^j} S_j(x_i[m], \mathbf{pa}_{M_j}[m]). \qquad (2)$$

For example, in the case of multinomial table CPTs, we have one sufficient statistic function for each joint assignment $x \in Val(\mathbf{M}_j), \mathbf{u} \in Val(\mathbf{Pa}_{M_j})$, which is $\eta\{X_i[m] = x, \mathbf{pa}_{M_j}[m] = \mathbf{u}\}$ — the indicator function that takes the value 1 if the event $(X_i[m] = x, \mathbf{Pa}_{M_j}[m] = \mathbf{u})$ holds, and 0 otherwise. The statistic on the data is

$$\hat{S}_j[x, \mathbf{u}] = \sum_{m=1}^{M} \sum_{X_i \in \mathbf{X}^j} \eta\{X_i[m] = x, \mathbf{Pa}_{M_j}[m] = \mathbf{u}\}$$

Given these sufficient statistics, the formula for the module likelihood function is:

$$L_j(\mathbf{Pa}_{M_j}, \mathbf{X}^j, \theta_{M_j \mid \mathbf{Pa}_{M_j}} : \mathcal{D}) = \prod_{x, \mathbf{u} \in Val(M_j, \mathbf{Pa}_{M_j})} \theta_{x \mid \mathbf{u}}^{\hat{S}_j[x, \mathbf{u}]}.$$

This term is precisely the one we would use in the likelihood of Bayesian networks. The only difference is that the vector of sufficient statistics for a local likelihood term is pooled over all the variables in the corresponding module. For example, consider the likelihood function for the module network of Figure 1(b). In this network we have three modules. The first consists of a single variable and has no parent, and so the vector of statistics $\hat{S}[\mathbf{M}_1]$ is the same as the statistics of the same variable $\hat{S}[MSFT]$. The second module contains three variables, and we have that the sufficient statistics for the module CPT is the sum of the statistics we would collect in the ground Bayesian network of Figure 1(a): $\hat{S}[\mathbf{M}_2, MSFT] = \hat{S}[AMAT, MSFT] + \hat{S}[MOT, MSFT] + \hat{S}[INTL, MSFT]$. Finally, $\hat{S}[\mathbf{M}_3, AMAT, INTL] = \hat{S}[DELL, AMAT, INTL] + \hat{S}[HPQ, AMAT, INTL]$.

As usual, the decomposition of the likelihood function allows us to perform maximum likelihood or MAP parameter estimation efficiently, optimizing the parameters for each module separately. The details are standard, and omitted for lack of space.

### 3.2 Priors and the Bayesian Score

As we discussed, our approach for learning module networks is based on the use of a Bayesian score. Specifically, we define a model score for a pair $(\mathcal{S}, \mathcal{A})$ as the posterior probability of the pair, integrating out the possible choices for the parameters $\theta$. We define an assignment prior $P(\mathcal{A})$, a structure prior $P(\mathcal{S} \mid \mathcal{A})$ and a parameter prior $P(\theta \mid \mathcal{S}, \mathcal{A})$. These describe our preferences over different networks *before* seeing the data. By Bayes' rule, we then have

$$P(\mathcal{S}, \mathcal{A} \mid \mathcal{D}) \propto P(\mathcal{A})P(\mathcal{S} \mid \mathcal{A})P(\mathcal{D} \mid \mathcal{S}, \mathcal{A})$$

where the last term is the *marginal likelihood*

$$P(\mathcal{D} \mid \mathcal{S}, \mathcal{A}) = \int P(\mathcal{D} \mid \mathcal{S}, \mathcal{A}, \theta)P(\theta \mid \mathcal{S})d\theta.$$

We define the Bayesian score as the log of $P(\mathcal{S}, \mathcal{A} \mid \mathcal{D})$, ignoring the normalization constant

$$\begin{aligned} \text{score}(\mathcal{S}, \mathcal{A} : \mathcal{D}) = & \qquad (3) \\ \log P(\mathcal{A}) + \log P(\mathcal{S} \mid \mathcal{A}) + \log P(\mathcal{D} \mid \mathcal{S}, \mathcal{A}) \end{aligned}$$

The main question is how to evaluate the score for different choices of $\mathcal{A}$ and $\mathcal{S}$. As we are going to examine a large number of alternatives, we need to be able to do this efficiently. In the case of Bayesian network learning, we can perform this task efficiently when the priors satisfy certain conditions. The same general ideas carry over to module networks, and so we review them briefly.

**Definition 3.1:** Let $P(\mathcal{A})$, $P(\mathcal{S} \mid \mathcal{A})$, $P(\theta \mid \mathcal{S}, \mathcal{A})$ be assignment, structure, and parameter priors.

- $P(\theta \mid \mathcal{S}, \mathcal{A})$ satisfies *parameter independence* if

$$P(\theta \mid \mathcal{S}, \mathcal{A}) = \prod_{j=1}^{K} P(\theta_{M_j \mid \mathbf{Pa}_{M_j}} \mid \mathcal{S}, \mathcal{A}).$$

- $P(\theta \mid \mathcal{S}, \mathcal{A})$ satisfies *parameter modularity* if $P(\theta_{M_j \mid \mathbf{Pa}_{M_j}} \mid \mathcal{S}_1, \mathcal{A}) = P(\theta_{M_j \mid \mathbf{Pa}_{M_j}} \mid \mathcal{S}_2, \mathcal{A})$ for all structures $\mathcal{S}_1$ and $\mathcal{S}_2$ such that $\mathbf{Pa}_{M_j}^{\mathcal{S}_1} = \mathbf{Pa}_{M_j}^{\mathcal{S}_2}$.

- $P(\theta, \mathcal{S} \mid \mathcal{A})$ satisfies *assignment independence* if $P(\theta \mid \mathcal{S}, \mathcal{A}) = P(\theta \mid \mathcal{S})$ and $P(\mathcal{S} \mid \mathcal{A}) = P(\mathcal{S})$.

- $P(\mathcal{S})$ satisfies *structure modularity* if $P(\mathcal{S}) \propto \prod_j \rho_j(\mathcal{S}_j)$ where $\mathcal{S}_j$ denotes the choice of parents for module $\mathbf{M}_j$, and $\rho_j$ is a distribution over the possible parent sets for module $\mathbf{M}_j$.

- $P(\mathcal{A})$ satisfies *assignment modularity* if $P(\mathcal{A}) \propto \prod_j \alpha_j(\mathcal{A}_j)$, where $\mathcal{A}_j$ is the choice of variables assigned to module $\mathbf{M}_j$, and $\{\alpha_j : j = 1, \dots, K\}$ is a family of functions from $2^{\mathcal{X}}$ to the positive reals. ∎

Parameter independence, parameter modularity, and structure modularity are the natural analogues of standard assumptions in Bayesian network learning [14]. Parameter independence implies that $P(\theta \mid \mathcal{S}, \mathcal{A})$ is a product of terms that parallels the decomposition of the likelihood in Eq. (1), with one prior term per local likelihood term $L_j$.

Parameter modularity states that the prior for the parameters of a module $M_j$ depends only on the choice of parents for $M_j$ and not on other aspects of the structure. Structure modularity implies that the prior over the structure $\mathcal{S}$ is a product of terms, one per each module.

Two assumptions are new to module networks. Assignment independence makes the priors on the parents and parameters of a module independent of the exact set of variables assigned to the module. Assignment modularity implies that the prior on $\mathcal{A}$ is proportional to a product of local terms, one corresponding to each module. Thus, the reassignment of one variable from one module $M_i$ to another $M_j$ does not change our preferences on the assignment of variables in modules other than $i, j$.

As for the standard conditions on Bayesian network priors, the conditions we define are not universally justified, and one can easily construct examples where we would want to relax them. However, they simplify many of the computations significantly, and are therefore very useful even if they are only a rough approximation. Moreover, the assumptions, although restrictive, still allow broad flexibility in our choice of priors. For example, we can encode preference (or restrictions) on the assignments of particular variables to specific modules. In addition, we can also encode preference for particular module sizes.

When the priors satisfy the assumptions of Definition 3.1, the Bayesian score decomposes into local *module scores*:

$$\text{score}(\mathcal{S}, \mathcal{A} : \mathcal{D}) = \sum_{j=1}^{K} \text{score}_{M_j}(\mathbf{Pa}_{M_j}, \mathcal{A}(\mathbf{X}^j) : \mathcal{D})$$

$$\text{score}_{M_j}(\mathbf{U}, \mathbf{X} : \mathcal{D}) =$$

$$\log \int L_j(\mathbf{U}, \mathbf{X}, \theta_{M_j | \mathbf{U}} : \mathcal{D}) P(\theta_{M_j} \mid S_j = \mathbf{U})$$

$$+ \log P(S_j = \mathbf{U}) + \log P(\mathcal{A}_j = \mathbf{X}) \qquad (4)$$

where $S_j = \mathbf{U}$ denotes that we chose a structure where $\mathbf{U}$ are the parents of module $M_j$, and $\mathcal{A}_j = \mathbf{X}$ denotes that $\mathcal{A}$ is such that $\mathbf{X}^j = \mathbf{X}$. As we shall see below, this decomposition plays a crucial rule in our ability to devise an efficient learning algorithm that searches the space of module networks for one with high score.

The only question is how to evaluate the integral over $\theta_{M_j}$ in $\text{score}_{M_j}(\mathbf{U}, \mathbf{X} : \mathcal{D})$. This depends on the parametric forms of the CPT and the form of the prior $P(\theta_{M_j} \mid S)$. Usually, we choose priors that are *conjugate* to the parameter distributions. Such a choice often leads to closed form analytic formula of the value of the integral as a function of the sufficient statistics of $L_j(\mathbf{Pa}_{M_j}, \mathbf{X}^j, \theta_{M_j | \mathbf{Pa}_{M_j}} : \mathcal{D})$. The details are standard [13] and omitted for lack of space.

## 4 Learning Algorithm

Given a scoring function over networks, we now consider how to find a high scoring module network. This problem is a challenging one, as it involves searching over two combinatorial spaces simultaneously — the space of structures and the space of module assignments. We therefore simplify our task by using an iterative approach that repeats two steps: In one step, we optimize a dependency structure relative to our current assignment function, and in the other, we optimize an assignment function relative to our current dependency structure.

**Structure Search Step.** The first type of step in our iterative algorithm learns the structure $\mathcal{S}$, assuming that $\mathcal{A}$ is fixed. This step involves a search over the space of dependency structures, attempting to maximize the score defined in Eq. (3). This problem is analogous to the problem of structure learning in Bayesian networks. We use a standard heuristic search over the combinatorial space of dependency structures. We define a search space, where each state in the space is a legal parent structure, and a set of operators that take us from one state to another. We traverse this space looking for high scoring structures using a search algorithm such as greedy hill climbing.

In many cases, an obvious choice of local search operators involves steps of adding or removing a variable $X_i$ from a parent set $\mathbf{Pa}_{M_j}$. (Note that edge reversal is not a well-defined operator for module networks, as an edge from a variable to a module represents a one-to-many relation between the variable and all of the variables in the module.) When an operator causes a parent $X_i$ to be added to a module $M_j$, we need to verify that the resulting module graph remains acyclic, relative to the current assignment $\mathcal{A}$. Note that this step is quite efficient, as cyclicity is tested on the module graph, which contains only $K$ nodes, rather than on the dependency graph of the ground Bayesian network, which contains $n$ nodes (usually $n \gg K$).

Also note that, as in Bayesian networks, the decomposition of the score provides considerable computational savings. When updating the dependency structure for a module $M_j$, the module score for another module $M_k$ does not change, nor do the changes in score induced by various operators applied to the dependency structure of $M_k$. Hence, after applying an operator to $\mathbf{Pa}_{M_j}$, we need only update the delta score for those operators that involve $M_j$.

**Module Assignment Search Step.** The second type of step in our iteration learns a new assignment function $\mathcal{A}$ from data, assuming that the module network structure $\mathcal{S}$ is given. Specifically, given a fixed structure $\mathcal{S}$ we want to find $\mathcal{A} = \text{argmax}_{\mathcal{A}'} \text{score}_M(\mathcal{S}, \mathcal{A}' : \mathcal{D})$.

Naively, we might think that we can further decompose the score across variables, allowing us to determine independently the optimal assignment $\mathcal{A}(X_i)$ for each variable $X_i$. Unfortunately, this is not the case. Most obviously, the assignments to different variables must be constrained so that the module graph remains acyclic. For example, if $X_1 \in \mathbf{Pa}_{M_i}$ and $X_2 \in \mathbf{Pa}_{M_j}$, we cannot simultaneously assign $\mathcal{A}(X_1) = j$ and $\mathcal{A}(X_2) = i$. More subtly, the Bayesian score for each module depends non-additively on the sufficient statistics of all the variables assigned to the module. (The log-likelihood function is additive in the sufficient statistics of the different variables, but the log marginal likelihood is not.) Thus, we can only compute the delta score for moving a variable from one module to another given a *fixed* assignment of the other variables to these two modules.

We therefore use a sequential update algorithm that reassigns the variables to modules one by one. The idea is simple. We start with an initial assignment function $\mathcal{A}^0$, and in a "round-robin" fashion iterate over all of the variables one at a time, and consider changing their module assignment. When considering a reassignment for a variable $X_i$, we keep the assignments of all other variables fixed and find the optimal legal (acyclic) assignment for $X_i$ relative to the fixed assignment. We continue reassigning variables until no single reassignment can improve the score.

The key to the correctness of this algorithm is its sequential nature: Each time a variable assignment changes, the assignment function as well as the associated sufficient statistics are updated before evaluating another variable. Thus, each change made to the assignment function leads to a legal assignment which improves the score. Our algorithm terminates when it can no longer improve the score. Hence, it converges to a local maximum, in the sense that no single assignment change can improve the score.

The computation of the score is the most expensive step in the sequential algorithm. Once again, the decomposition of the score plays a key role in reducing the complexity of this computation: When reassigning a variable $X_i$ from one module $\mathbf{M}_j$ to another $\mathbf{M}_k$, only the local score of these modules changes.

**Convergence.** Our algorithm starts with an initial guess of assignment (see below), and then applies the two steps described above iteratively until convergence. We have constructed our iterative algorithm so that each of the two steps — structure update and assignment update — is guaranteed to either improve the score or leave it unchanged.

**Theorem 4.1:** The iterative module network learning algorithm converges to a local maximum of $\mathrm{score}(\mathcal{S}, \mathcal{A} : \mathcal{D})$.

**Initialization.** The only remaining question is how to choose the initial module assignment to begin the iterative algorithm. Recall that we need to find a way to group variables into initial modules. Ideally, this initialization would put together variables that behaved similarly in the different instances. This problem can be thought of as a clustering problem, where the objects to be clustered are the variables in the module network and their features are their behavior in the different instances in the original data set. For example, in our stock market example, we would cluster stocks based on the similarity of their behavior over different trading days. (Note that, when viewing the data from the perspective of learning a Bayesian network or a module network, the "instances" are trading days and their attributes are stocks.) We can use any standard clustering procedure (e.g., [2]) to come up with this initial clustering.

We choose to use a procedure that is suitable to our problem, in that it evaluates a partition of variables into modules by measuring the extent to which the module model is a good fit to the data of the variables in the module. This algorithm can be best thought of as performing *model merging* (as in [7]), in a module network with a specific structure. However, instead of merging values of random variables, we merge modules. We start by building a module network as follows. We introduce a dummy variable $U$ that encodes training instance identity — $u[m] = m$ for all $m$. We then create $n$ modules, with $\mathcal{A}(X_i) = i$, and $\mathbf{Pa}_{\mathbf{M}_i} = U$. Note that, in this network, each instance and each variable has its own local probabilistic model.

Next, we consider all possible legal module mergers (those corresponding to modules with the same domain), where we change the assignment function to replace two modules $j_1$ and $j_2$ by a new module $j_{1,2}$. Note that, following the merger, each instance still has a different probabilistic model, but the two variables $X_{j_1}$ and $X_{j_2}$ now must share parameters. We evaluate each such merger by computing the score of the resulting module network. We then greedily choose the merger that leads to the best scoring network. Thus, the procedure will merge two modules that are similar to each other across the different instances. We continue to do these mergers until we reach a module network with the desired number of modules, as specified in the original choice of $\mathcal{C}$.

## 5 Learning with Regression Trees

We now briefly review the conditional distribution we use in the experiments below. Many of the domains suited for module network models contain continuous valued variables, such as gene expression or price changes in the stock market. For these domains, we often use a conditional probability model represented as a *regression tree* [1]. For our purposes, a regression tree $T$ for $P(X \mid \mathbf{U})$ is defined via a rooted binary tree, where each *node* in the tree is either a *leaf* or an *interior node*. Each interior node is labeled with a test $U < u$ on some variable $U \in \mathbf{U}$ and $u \in \mathbb{R}$. Such an interior node has two outgoing *arcs* to its children, corresponding to the outcomes of the test (true or false). The tree structure $T$ captures the *local* dependency structure of the conditional distribution. The parameters of $T$ are the distributions associated with each leaf. In our implementation, each leaf $\ell$ is associated with a univariate Gaussian distribution over values of $X$, parameterized by a mean $\mu_\ell$ and variance $\sigma_\ell^2$.

To learn module networks with regression-tree CPTs, we must extend our previous discussion by adding another component to $\mathcal{S}$ that represents the trees $T_1, \ldots, T_K$ associated with the different modules. Once we specify these components, the above discussion applies with several small differences. These issues are similar to those encountered when introducing decision trees to Bayesian networks [4, 9], and so we only briefly touch on them.

Given a regression tree $T_j$ for $P(\mathbf{M}_j \mid \mathbf{Pa}_{\mathbf{M}_j})$, the corresponding sufficient statistics are the statistics of the distributions at the leaves of the tree. For each leaf $\ell$ in the tree, and for each data instance $\mathbf{x}[m]$, we let $\ell_j[m]$ denote the leaf reached in the tree given the assignment to $\mathbf{Pa}_{\mathbf{M}_j}$ in $\mathbf{x}[m]$. The module likelihood decomposes as a product of terms, one for each leaf $\ell$. Each term is the likelihood for the Gaussian distribution $\mathcal{N}\left(\mu_\ell; \sigma_\ell^2\right)$, with the usual sufficient statistics for a Gaussian distribution.

When performing structure search for module networks with regression-tree CPTs, in addition to choosing the parents of each module, we must also choose the associated tree structure. We use the search strategy proposed in [4], where the search operators are leaf splits. Such a *split* operator replaces a leaf in a tree $T_j$ with an internal node with some test on a variable $U$. The two branches below the newly created internal node point to two new leaves, each with its associated Gaussian. This operator must check for acyclicity, as it implicitly adds $U$ as a parent of $\mathbf{M}_j$. When performing the search, we consider splitting each possible leaf on each possible parent $U$ and each value $u$. As always in regression-tree learning, we do not have to consider all real values $u$ as possible split points; it suffices to consider values that arise in the data set.

## 6    Experimental Results

We evaluated our module network learning procedure on synthetic data and on two real data sets — gene expression data, and stock market data. In all cases, our data consisted solely of continuous values. As all of the variables have the same domain, the definition of the module set reduces simply to a specification of the total number of modules. We used regression trees as the local probability model for all modules. As our search algorithm, we used beam search, using a lookahead of three splits to evaluate each operator. When learning Bayesian networks, as a comparison, we used precisely the same structure learning algorithm, simply treating each variable as its own module.

**Synthetic data.** As a basic test of our procedure in a controlled setting, we used synthetic data generated by a known module network. This gives a known ground truth to which we can compare the learned models. To make the data realistic, we generated synthetic data from a model that was learned from the gene expression dataset described below. The generating model had 10 modules and a total of 35 variables that were a parent of some module. From the learned module network, we selected 500 variables, including the 35 parents. We tested our algorithm's ability to reconstruct the network using different numbers of modules; this procedure was run for training sets of various sizes ranging from 25 instances to 500 instances, each repeated 10 times for different training sets.

We first evaluated the generalization to unseen test data, measuring the likelihood ascribed by the learned model to 4500 unseen instances. The results, summarized in Figure 2(a), show that, for all training set sizes, except the smallest one with 25 instances, the model with 10 modules performs the best. As expected, models learned with larger training sets do better; but, when run using the correct number of 10 modules, the gain of increasing the number of data instances beyond 100 samples is small.

A closer examination of the learned models reveals that, in many cases, they are almost a 10-module network. As shown in Figure 2(b), models learned using 100, 200, or 500 instances and up to 50 modules assigned $\geq 80\%$ of the variables to 10 modules. Indeed, these models achieved high performance in Figure 2(a). However, models learned with a larger number of modules had a wider spread for the assignments of variables to modules and consequently achieved poor performance.

Finally, we evaluated the model's ability to recover the correct dependencies. The total number of parent-child relationships in the generating model was 2250. For each model learned, we report the fraction of correct parent-child relationships it contains. As shown in Figure 2(c), our procedure recovers 74% of the true relationships when learning from a dataset of size 500 instances. Once again, we see that, as the variables begin fragmenting over a large number of modules, the learned structure contains many spurious relationships. Thus, our results suggest that, in domains with a modular structure, statistical noise is likely to prevent overly detailed learned models such as Bayesian networks from extracting the commonality between different variables with a shared behavior.

**Gene Expression Data.**    We next evaluated the performance of our method on a real world data set of gene expression measurements. A *microarray* measures the activity level (mRNA expression level) of thousands of genes in the cell in a particular condition. We view each experiment as an instance, and the expression level of each measured gene as a variable [10]. In many cases, the coordinated activity of a group of genes is controlled by a small set of *regulators*, that are themselves encoded by genes. Thus, the activity level of a regulator gene can often predict the activity of the genes in the group. Our goal is to discover these modules of co-regulated genes, and their regulators.

We used the expression data of [11], which measured the response of yeast to different stress conditions. The data consists of 6157 genes and 173 experiments. In this domain, we have prior knowledge of which genes are likely to play a regulatory role. Subsequently, we restricted the possible parents to 466 yeast genes that may play such a role. We then selected 2355 genes that varied significantly in the data and learned a module network over these genes. We also learned a Bayesian network over this data set.

We evaluated the generalization ability of different models, in terms of log-likelihood of test data, using 10-fold cross validation. In Figure 3(a), we show the difference between module networks of different size and the baseline Bayesian network, demonstrating that module networks generalize much better to unseen data for almost all choices of number of modules.

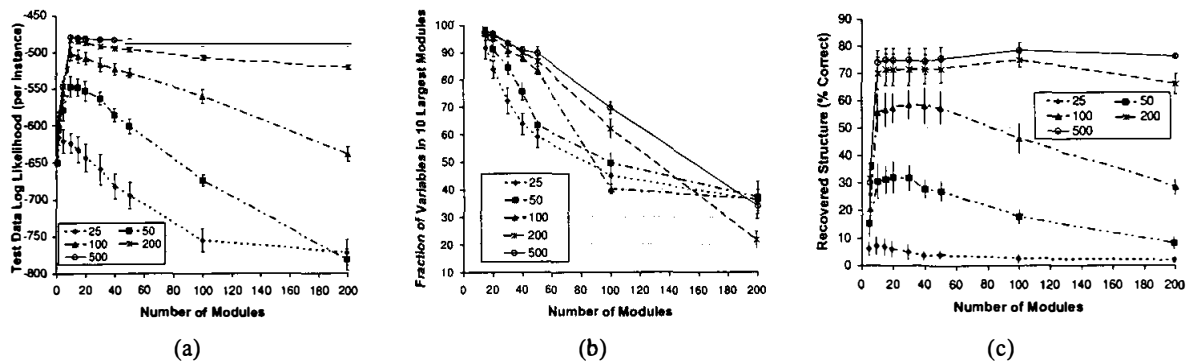

Figure 2: Performance of learning from synthetic data as a function of the number of modules and training set size. In all cases, the x-axis corresponds to the number of modules, each curve corresponds to a different number of training instances, and each point shows the mean and standard deviations from the 10 sampled data sets. (a) Log-likelihood per instance assigned to held-out data. (b) Fraction of variables assigned to the largest 10 modules. (c) Average percentage of correct parent-child relationships recovered.

We next tested the biological validity of the learned module network with 50 modules. (We selected 50 modules due to the biological plausibility of having, on average, 40–50 genes per module.) First, we examined whether genes in the same module have shared functional characteristics. To this end, we used annotations of the genes' biological functions from the Saccharomyces Genome Database [3]. We systematically evaluated each module's gene set by testing for significantly enriched annotations. Suppose we find $l$ genes with a certain annoation in a module of size $N$. To check for enrichment, we calculate the *p-value* of these numbers — the probability of finding that many genes of that annotation in a random subset of $N$ genes. For example, the "protein folding" module contains 10 genes, 7 of which are annotated as protein folding genes. In the whole data set, there are only 26 genes with this annotation. Thus, the p-value of this annotation, that is, the probability of choosing 7 or more genes in this category by choosing 10 random genes, is less than $10^{-12}$. Our evaluation showed that 42 (resp. 20) modules, out of 50, had at least one significantly enriched annotation with a p-value less than 0.005 (resp. less than $10^{-6}$). Furthermore, the enriched annotations reflect the key biological processes expected in our dataset. We used these annotations to label the modules with meaningful biological names.

We can use these annotations to reason about the dependencies between different biological processes at the module level. For example, we find that the *cell cycle* module, regulates the *histone* module. The cell cycle is the process in which the cell replicates its DNA and divides, and it is indeed known to regulate histones — key proteins in charge of maintaining and controlling the DNA structure. Another module regulated by the cell cycle module is the *nitrogen catabolite repression (NCR)* module, a cellular response activated when nitrogen sources are scarce. We find that the *NCR* module regulates the *amino acid metabolism, purine metabolism* and *protein synthesis* modules, all representing nitrogen-requiring processes, and hence likely to be regulated by the *NCR* module. These examples demonstrate the insights that can be gleaned from a higher order model, and which would have been obscured in the unrolled Bayesian network over 2355 genes.

**Stock Market Data.** In a very different application, we examined a data set of NASDAQ stock prices. We collected stock prices for 2143 companies, in the period 1/1/2002–2/3/2003, covering 273 trading days. We took each stock to be a variable, and each instance to correspond to a trading day, where the value of the variable is the log of the ratio between that day's and the previous day's closing stock price. This choice of data representation focuses on the relative changes to the stock price, and eliminates the magnitude of the price itself (which depends on such irrelevant factors as the number of outstanding shares). As potential controllers, we selected 250 of the 2143 stocks, whose average trading volume was the largest across the dataset.

As with gene expression data, we used cross validation to evaluate the generalization ability of different models. As we can see in Figure 3(b), module networks perform significantly better than Bayesian networks in this domain.

To test the quality of our modules, we measured the enrichment of the modules in the network with 50 modules for annotations representing various sectors to which each stock belongs. We found significant enrichment for 21 such annotations, covering a wide variety of sectors. We also compared these results to the clusters of stocks obtained from applying Autoclass [2] to the data. Here, as we described above, each instance corresponds to a stock and is described by 273 random variables, each representing a trading day. In 20 of the 21 cases, the enrichment was far more significant in the modules learned using module networks compared to the one learned by AutoClass, as can be seen in Figure 3(c).

Finally, we also looked at the structure of the module network, and found several cases where the structure fit our understanding of the stock domain. Several modules corresponded primarily to high tech stocks. One of these, consisting mostly of software, semi-conductor, communication, and broadcasting services, had as its two main predictors Molex, a large manufacturer of electronic, electrical and fiber optic interconnection products and systems, and Atmel, specializing in design, manufacturing and marketing of advanced semiconductors. Molex was also the parent for another module, consisting primarily of software, semiconductor, and medical equipment companies; this module had as additional parents Maxim, which develop integrated circuits, and Affymetrix, which designs and develops gene microarray chips. In this, as in many other cases, the parents of a module are from similar sectors as the stocks in the module.

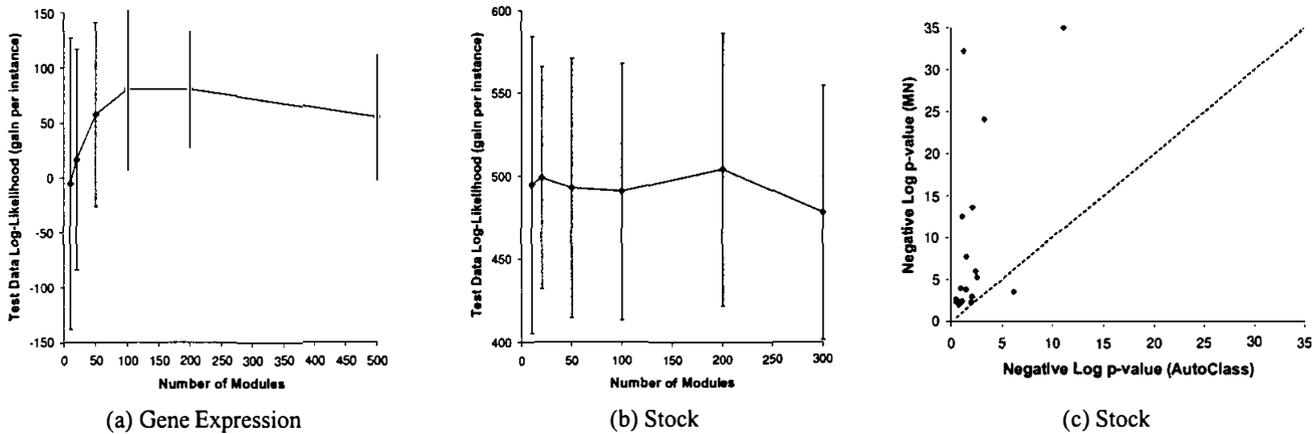

|  |  |  |
|---|---|---|
| (a) Gene Expression | (b) Stock | (c) Stock |

Figure 3: (a) & (b) Comparison of generalization ability of module networks learning with different numbers of modules on the gene expression and stock data sets. The x-axis denotes the number of modules. The y-axis denotes the difference in log-likelihood on held out data between the learned module network and the learned Bayesian network, averaged over 10 folds; the error bars show the standard deviation. (c) Comparison of the enrichment for annotations of sectors between the modules learned using module networks and the clusters used from by applying [2]. Each point corresponds to an annotation, and the x and y axes are the negative log p-values of its enrichment for the two models.

## 7 Discussion and Conclusions

We have introduced the framework of *module networks*, an extension of Bayesian networks that includes an explicit representation of *modules* — subsets of variables that share a statistical model. We have presented a Bayesian learning framework for module networks, that learns both the partitioning of variables into modules and the dependency structure of each module. We showed experimental results on two complex real-world data sets, each including measurements of thousands of variables, in the domains of gene expression and stock market. Our results show that our learned module networks have much higher generalization performance than a Bayesian network learned from the same data.

There are several reasons why a learned module network is a better model than a learned Bayesian network. Most obviously, parameter sharing between variables in the same module allows each parameter to be estimated based on a much larger sample. Moreover, this allows us to learn dependencies that are considered too weak based on statistics of single variables. These are well-known advantages of parameter sharing; the interesting aspect of our method is that we determine automatically which variables have shared parameters.

More interestingly, the assumption of shared structure significantly restricts the space of possible dependency structures, allowing us to learn more robust models than those learned in a classical Bayesian network setting. While the variables in the same module might behave according to the same model in underlying distribution, this will often not be the case in the empirical distribution based on a finite number of samples. A Bayesian network learning algorithm will treat each variable separately, optimizing the parent set and CPD for each variable in an independent manner. In the very high-dimensional domains in which we are interested, there are bound to be spurious correlations that arise from sampling noise, inducing the algorithm to choose parent sets that do not reflect real dependencies, and will not generalize to unseen data. Conversely, in a module network setting, a spurious correlation would have to arise between a possible parent and a large number of other variables before the algorithm would find it worthwhile to introduce the dependency.

Module networks are related both to the framework of *object-oriented Bayesian networks* (OOBNs) [15] and to the framework of *probabilitic relational models* (PRMs) [16, 8]. These frameworks extend Bayesian Networks to a setting involving multiple related objects, and allow random variables of the same *class* to share parameters and dependency structure. In the module network framework, we can view each variable as an object and each module as a class, so that the variables in a single module share the same probabilistic model. As the module assignments are not known in advance, module networks correspond most closely to the variant of these frameworks where there is *type uncertainty* — uncertainty about the class assignment of objects. However, despite this high-level similarity, the module network framework differs in certain key points from both OOBNs and PRMs, with significant impact on the learning task.

In OOBNs, objects in the same class must have the same internal structure and parameterization, but can depend on different sets of variables (as specified in the mapping of variables in an object's interface to its actual inputs). By contrast, in a module network, all of the variables in a module (class) must have the same specific parents. This assumption greatly reduces the size and complexity of the hypothesis space, leading to a more robust learning algorithm. On the other hand, this assumption requires that we be careful in making certain steps in the structure search, as they have more global effects than on just one or two variables. Due to these differences, we cannot simply apply an OOBN structure-learning algorithm, such as the one proposed by Langseth and Nielsen [18], to such complex, high-dimensional domains.

In PRMs, the probabilistic dependency structure of the objects in a class is determined by the relational structure of the domain (e.g., the *Cost* attribute of a particular car object might depend on the *Income* attribute of the object representing this particular car's owner). In the case of module networks, there is no known relational structure to which probabilistic dependencies can be attached. Without such a relational structure, PRMs only allow dependency models specified at the class level. Thus, we can assert that the objects in one class depend on some aggregate quantity of the objects in another. We cannot, however, state a dependence on a particular object in the other class (without some relationship specified in the model). Getoor *et al.* [12]) attempt to address this issue using a class hierarchy . Their approach is very different from ours, requiring some fairly complex search steps, and is not easily applied to the types of domains considered in this paper. Overall, module networks do not apply as broadly as PRMs, but allow much more flexible parameter sharing and dependency structures in domains where they apply.

There are several important extensions to the work we presented here. Most obviously, we have not addressed the issue of selecting the number of modules. We can adapt Bayesian scoring criteria used to evaluate standard clustering methods [2, 7] for the problem of evaluating different choices for the number of modules. However, much remains to be done on the problem of proposing new modules and initializing them.

In this paper, we focused on the statistical properties of our method. In a companion biological paper [19], we use the module network learned from the gene expression data described above to predict gene regulation relationships. There, we performed a comprehensive evaluation of the validity of the biological structures reconstructed by our method. By analyzing biological databases and previous experimental results in the literature, we confirmed that many of the regulatory relations that our method automatically inferred are indeed correct. Furthermore, our model provided focused predictions for genes of previously uncharacterized function. We performed wet lab biological experiments that confirmed the 3 novel predictions we tested. Thus, we have demonstrated that the module network model is robust enough to learn a good approximation of the dependency structure between 2355 genes using only 173 samples. These results show that, by learning a structured probabilistic representation, we identify regulation networks from gene expression data and successfully address one of the centeral problems in analysis of gene expression data.

**Acknowledgements.**    We thank the anonymous reviewers for useful comments on a previous version of this paper. E. Segal, D. Koller, and N. Friedman were supported in part by NSF grant ACI-0082554 under the ITR Program. E. Segal was also supported by a Stanford Graduate Fellowship (SGF). A. Regev was supported by the Colton Foundation. D. Pe'er was supported by an Eshkol Fellowship. N. Friedman was also supported by an Alon Fellowship, by the Harry & Abe Sherman Senior Lectureship in Computer Science, and by the Israeli Ministry of Science.